\documentclass[lettersize,journal]{IEEEtran}
\usepackage{amsmath,amsfonts, xcolor}
\usepackage{algorithm}
\usepackage{algpseudocode}

\usepackage{array}
\usepackage[caption=false,font=normalsize,labelfont=sf,textfont=sf]{subfig}
\usepackage{textcomp}
\usepackage{stfloats}
\usepackage{url}
\usepackage{verbatim}
\usepackage{graphicx}
\usepackage{cite}
\usepackage{makecell}

\usepackage{multirow, color, rotating, subcaption}
\usepackage{hyperref}
\begin{document}

\title{A Hierarchical IDS for Zero-Day Attack Detection in Internet of Medical Things Networks}

\author{Md Ashraf Uddin, Nam H. Chu, and Reza Rafeh

\thanks{The authors are with the School of Information Technology, Crown Institute of Higher Education, Australia. (emails: \{\textit{ashraf.uddin, namhoai.chu, reza.rafeh\}@cihe.edu.au})}
\thanks{Manuscript received XX XX, 2024; revised XX YY, 2025.}}

\markboth{Journal of \LaTeX\ Class Files,~Vol.~XX, No.~XX, XX~2025}%
{Shell \MakeLowercase{\textit{et al.}}: A Sample Article Using IEEEtran.cls for IEEE Journals}


\maketitle

\begin{abstract}
    The Internet of Medical Things (IoMT) has been emerging as the main driver for the healthcare revolution. 
        These networks typically include resource-constrained, heterogeneous devices such as wearable sensors, smart pills, and implantable devices, making them vulnerable to diverse cyberattacks, e.g., denial-of-service, ransomware, data hijacking, and spoofing attacks. 
        To mitigate these risks, Intrusion Detection Systems (IDSs) are critical for monitoring and securing patients' medical devices. 
        However, traditional centralized IDSs may not be suitable for IoMT due to inherent limitations such as delays in response time, privacy concerns, and increased security vulnerabilities. 
        Specifically, centralized IDS architectures require every sensor to transmit its data to a central server, 
        potentially causing significant delays or even disrupting network operations in densely populated areas. 
        On the other hand, executing an IDS on IoMT devices is generally infeasible due to the lack of computational capacity. 
        Even if some lightweight IDS components 
        can be deployed in these devices, they must wait for the centralized IDS to provide updated models, otherwise, they will be vulnerable to zero-day attacks, posing significant risks to patient health and data security. 
        To address these challenges, we propose a novel multi-level IoMT IDS framework that can not only detect zero-day attacks but also differentiate between known and unknown attacks. 
        In particular, the first layer, namely the near Edge, filters network traffic at coarse level (i.e., attack or not), by leveraging 
        meta-learning or One Class Classification (OCC) based on the usfAD algorithm. 
        Then, the deeper layers (e.g., far Edge and Cloud) will determine whether the attack is known or unknown, as well as the detailed type of attack. 
        The experimental results on the latest IoMT dataset CICIoMT2024 show that our proposed solution achieves high performance, i.e., 99.77\% accuracy and 97.8\% F1-score.
        Notably, the first layer, using either meta-learning or usfAD-based OCC, can detect zero-day attacks with high accuracy without requiring new datasets of these attacks, making our approach highly applicable for the IoMT environment.
         Furthermore, the meta-learning approach requires less than 1\% of the dataset to achieve high performance in attack detection.
    
\end{abstract}


\section{Introduction}
\label{section:introduction}
\IEEEPARstart{T}{he} Internet of Things (IoT) represents a transformative concept where interconnected devices equipped with sensors collect, analyze, and interact with the physical environment, creating networks that serve diverse applications. As the IoT evolves and expands into many areas in our daily lives, the Internet of Medical Things (IoMT) has emerged as a notable application, expected to revolutionize healthcare through remote diagnosis, patient monitoring, and enhanced treatment capabilities. 
    The IoMT integrates various device types, from wearable and implantable devices, smartphones, to cloud-based systems to enable continuous monitoring and management of medical conditions, offering patients improved care and convenience. 
    According to the report from the Knowledge Based Value's report  \cite{KBV2023}, the IoMT market is predicted to exceed 20.4\% compound annual growth rate (CAGR) and reach \$588.9 billion by 2030.

While IoMT provides significant healthcare advancements, the rapid expansion of IoMT also introduces significant cybersecurity challenges due to vulnerabilities in device security, data transmission, and storage, posing risks to patient safety and healthcare infrastructure. 
    Specifically, since IoMT networks comprise resource-constrained devices, e.g., wearable sensors, smartphones, and personal digital assistants (PDAs), that continuously collect and transmit patient data for analysis and diagnosis \cite{qadri2020future}, they are susceptible to a range of cyberattacks, including denial-of-service (DoS), ransomware, data hijacking, spoofing, and social engineering \cite{da2019internet}.
    It is reported that ransomware attacks in the US alone incur \$21 billion burden annually, and 21\% of ransomware attacks in the world target IoT/IoMT devices \cite{Levelblue2022}.
    Addressing these challenges requires robust security mechanisms and compliance with data protection frameworks such as HIPAA and GDPR to ensure the safety, privacy, and efficiency of IoMT systems \cite{wani2024security}.
    Moreover, the sensitive nature of medical data and the critical functionalities of IoMT devices magnify the potential risks, making robust cybersecurity mechanisms a necessity \cite{kasinathan2013denial}.
    
In this context, Intrusion Detection Systems (IDSs) based on machine learning (ML) are a key line of defense in safeguarding IoMT networks from malicious activities. 
    However, traditional centralized IDS architectures may be unsuitable for IoMT considering the distributed nature of this network. 
    Firstly, these systems require transmitting all sensor data to a central server, which introduces privacy concerns, increases response time, and is inefficient for resource-constrained IoMT devices \cite{butun2019security}. 
    Additionally, centralized processing creates single points of failure and network congestion bottlenecks, which are particularly problematic in time-critical medical scenarios. 
    Secondly, such architectures often rely heavily on large datasets for training machine learning models, which are infeasible to process locally on IoMT devices due to due to their memory constraints, processing power, and battery life limitations \cite{sarhan2023zero}. 
    Moreover, these systems are ill-equipped to handle zero-day attacks, which exploit unknown vulnerabilities and present significant risks to patient health and data security.

Given these challenges, there is a pressing need for an intrusion detection framework tailored for IoMT environments that can efficiently detect zero-day attacks while addressing privacy and computational constraints.      
    Medical devices are susceptible to cyberattacks, especially during communication with Near Edge devices, which themselves receive network traffic from Far Edge nodes. 
    To safeguard medical devices, it is essential to deploy IDS at the Near Edge nodes. 
    These systems can monitor and detect potential threats before they reach the medical devices. 
    While ideally, medical devices should also incorporate IDS, their resource constraints often make this impractical \cite{sarhan2023zero}. Therefore, both Near Edge and Far Edge devices should be equipped with IDS capabilities. However, Near Edge devices are also resource-limited and may not possess sufficient attack data to train traditional supervised machine learning or deep learning models effectively. To enable these devices to independently run IDS and train models without relying on a centralized system, a semi-supervised learning approach or lightweight supervised models requiring minimal training data should be considered. In addition, Near Edge nodes are the first line of defense and must be capable of detecting zero-day attacks. Traditional supervised models, while effective at recognizing known threats from historical data, are typically inadequate for identifying previously unseen attacks\cite{uddin2024usfad}. This highlights the need for more adaptive and data-efficient IDS solutions at the edge level.

To address these issues, this paper proposes a novel hierarchical intrusion detection system, leveraging meta learning and One Class Classification (OCC) based on the usfAD algorithm, to identify attacks without requiring extensive training datasets in the Near Edge. 
    Specifically, our hierarchical IDS enables medical end devices (i.e., Near Edge nodes) to locally identify network traffic as anomaly that might be known/historical or zero-day attacks.
    Then, Far Edge nodes can further determine identified threats as known or unknown attacks so that the admin can arrange training of model.
    Finally, Cloud nodes are able to classify specific historical or known attack categories as well as attack sub-categories. 
    By doing so. our distributed approach can enhance intrusion detection accuracy, ensure real-time response, and optimize resource utilization. 
    Our contributions are summarized below.

\begin{itemize}
    \item \textbf{Develop the hierarchical IDS framework tailored for IoMT networks}, consisting of multiple attack detection levels spanning from Near Edge and Far Edge nodes to the Cloud.
        By doing so, our proposed framework can leverages resources at different layers in IoMT simultaneously, thus mitigating the point-of-congestion problem in traditional flat IDSs. 
        
     \item \textbf{Propose Anomaly Detection at the Near Edge Layer} that enables medical end devices (such as smartphones and PDAs) to detect anomalous traffic locally by leveraging meta learning and usfAD-based OCC. 
     While OCC can effectively detect anomalous traffic by training models to recognize normal traffic patterns without attack samples, the meta-learning can achieve high detection accuracy with a small amount of data.      
     Since neither approach requires retraining or fine-tuning, our solution reduces computational overhead and eliminates frequent model updates, enhancing medical device practicality through real-time, on-device threat detection. 
    
     \item \textbf{Develop the Zero-day Attack Detection at the Far Edge Layer} that deploys OCC based on usfAD algorithm at the edge nodes to identify zero-day attacks, thus effectively balancing computational demands and real-time response while maintaining high detection accuracy. 
         
        
   \item \textbf{Evaluate the proposed solution comprehensively.} 
        Comprehensive experiments were conducted using the CICIoMT2024 dataset, a multi-modal protocol dataset designed for IoMT security solutions. 
        These experiments demonstrated the effectiveness of the proposed framework in detecting zero-day attacks, highlighting its superior performance compared to traditional learning methods.   
\end{itemize}


The remainder of this paper is organized as follows: Section \ref{Literature Review} discusses related work in the field of IoMT security and intrusion detection systems. 
    Section \ref{Proposed}  describes the proposed framework, including its architecture and learning algorithms. 
    Section \ref{RESULTS AND DISCUSSION} presents the experimental setup and results, followed by a detailed performance evaluation. 
    Finally, Section \ref{Conclusion} concludes the paper and outlines future research directions.

\section{Related works} 
\label{Literature Review}
With the explosion of IoMT, developing an adaptive and effective ML-based IDS for IoMT networks has attracted enormous attention from academia and industry since these systems often handle sensitive and life-critical medical traffic. 
    However, existing IoMT IDS solutions face several critical limitations that hinder their practical deployment. 
    First, regarding architectural design, most existing works consider centralized IDS architectures for IoMT networks \cite{rm2020effective, alrashdi2019fbad, kumar2021ensemble, khan2021hybrid, khan2022xsru, wang2024federated, dhanya2024novel, zhang2024comparative, kilincer2023automated, gupta2022tree,  fouda2022novel, alsalman2024comparative, saheed2021efficient, manimurugan2020effective, hady2020intrusion, ravi2023deep, khan2023secure, nandy2021intrusion, gao2017machine, he2019intrusion, newaz2020heka,  areia2024iomt, ioannou2024gemlids, sohail2024explainable, dadkhah2024ciciomt2024, mohammadi2024securing, ramesh2024efficient}.
    This centralized approach makes their solutions less suitable for IoMT networks due to resource limitations of IoMT devices as well as the point-of-congestion problem.
    Second, from an evaluation perspective, many works are evaluated on either non-IoMT datasets (e.g., NSL-KDD and ToN-IoT \cite{rm2020effective, alrashdi2019fbad, kumar2021ensemble, khan2021hybrid, khan2022xsru}) or simulated datasets (e.g., \cite{gao2017machine, he2019intrusion, newaz2020heka}). 
    Thus, these approaches may not achieve the same performance in practical IoMT networks, which often consist of heterogeneous resource-constrained devices.
    Third, concerning functionality limitations, many studies (e.g., \cite{wang2024federated, dhanya2024novel, zhang2024comparative, kumar2021ensemble, rm2020effective, kilincer2023automated, gupta2022tree, fouda2022novel, alsalman2024comparative, saheed2021efficient, hady2020intrusion, ravi2023deep, khan2023secure, nandy2021intrusion}) only consider binary classification, which determines whether traffic is attack or normal. 
    As such, these IDS solutions are unable to provide adequate information to help IoMT systems make further mitigation decisions.
    
To mitigate shortcomings of the centralized IDS architecture, other works consider a hierarchical/distributed IDS architectures for IoMT networks~\cite{begli2019layered, thamilarasu2020intrusion, gupta2022cybersecurity, singh2022dew}.
    Specifically, the work in \cite{begli2019layered} proposes a multi-layer IDS that deploys multiple agents.
    These agents are categorized into three layers: sensors, management, and databases.
    The first layer employs a Support Vector Machine (SVM)-based anomaly detector to examine the traffic to find potential attack.
    The second layer uses a signature-based IDS, aiming to further identify the type of attack.
    Then, the potential malicious traffic is examined at the third layer by a hybrid IDS that consists of an anomaly- and signature-based IDSs, attempting to correct the missed classification in the previous layers.
    The major weakness of this solution is that it was evaluated with the NSL-KDD dataset, which is not a IoMT dataset.
    Additionally, the second rule of the framework operation (i.e., ``If anomaly detection detects an attack and misuse detection does not detect any attack, then it is not an attack.'') makes the system unable to detect zero-day attacks.
    
Similarly, \cite{thamilarasu2020intrusion} proposes a mobile agent-based IDS that distributes agents across layers, i.e., sensor and cluster head.
    In this framework, sensor agents first identify a traffic as benevolent or malicious. 
    Then, a cluster head agent will share this information with other cluster head agents to make the decision based on a majority. 
    Here, sensor and cluster head agents use traditional machine learning algorithms (e.g., SVM, Decision Tree, and K-Nearest Neighbour). 
    The shortcomings of this study are using a simulated dataset for evaluation and deploying a binary classification for agents.
    Additionally, zero-day attacks are not considered.

Differently with~\cite{begli2019layered, thamilarasu2020intrusion}, the hierarchical deep learning-based IDS for IoMT networks is considered in \cite{gupta2022cybersecurity}. 
    In particular, each edge nodes maintain its local DL model, whose architecture is shallow (e.g., 2-3 layers) while the model in cloud is deeper (e.g., more than ten layers), aiming to achieve higher detection accuracy. 
    To speed up the training time of cloud models, the authors propose a method for merging and aggregating layers of trained edge model to build a partly pre-trained cloud model. 
    The main drawback of this work~\cite{gupta2022cybersecurity} is that both edge and cloud models only perform the binary classification task, i.e., attack or benign classes. 
    Additionally, although the framework is able detect zero-day attacks, it cannot differentiate whether attacks are known or unknown types.
Similar to~\cite{gupta2022cybersecurity}, the study  \cite{singh2022dew} develops a hierarchical deep learning-based IDS that leverages dew computing.
    Specifically, this framework deploys federated learning on distributed dew servers of the IoMT system, which can be referred to as Near Edge nodes in our work. 
    As this IDS considers binary classification and uses a dataset that not for IoMT (i.e., TON-IoT and NSL-KDD), this work shares the shortcomings with many above works, e.g.\cite{wang2024federated, dhanya2024novel, zhang2024comparative, kumar2021ensemble, rm2020effective, kilincer2023automated, gupta2022tree, fouda2022novel, alsalman2024comparative, saheed2021efficient, hady2020intrusion, ravi2023deep, khan2023secure, nandy2021intrusion, thamilarasu2020intrusion, gupta2022cybersecurity, begli2019layered}.

Recently, \cite{zukaib2024meta} proposes a hierarchical machine learning model for a IoMT IDS. 
    This model consists of multiple components, each leverages on machine learning algorithm, i.e., Decistion Tree (DT), Random Forest (RF), AdaBoost, Naive Bayes, Multilayer Perceptron (MLP), and XGBoost.
    Firstly, this framework uses the bat algorithm to optimize parameter of weak learners, including DT, RF, AdaBoost, NN, and MLP. 
    Subsequently, XGBoost is used as the meta-learner (which is more accurately an ensemble method) to produces the final ensemble prediction. 
    In fact, this study \cite{zukaib2024meta} has terminological problems, confusing readers. 
    Specifically, it uses ensemble learning/stacking, which is not meta-learning. 
    Meta-learning refers to ``learning to learn'' across multiple tasks (like MAML or Reptile that is leveraged in our work), not combining multiple models on a single task.
    Moreover, although detecting zero-day attacks is considered in this framework \cite{zukaib2024meta}, the detection performance of zero-day attacks is not conducted, making the evaluation incomplete.

\vspace{5 pt}
\noindent \textit{Research Gap and Challenges}

Given the above, existing approaches suffer from several critical limitations: (i) architectural inefficiencies that cannot scale to IoMT's distributed nature \cite{rm2020effective, alrashdi2019fbad, kumar2021ensemble, khan2021hybrid, khan2022xsru, wang2024federated, dhanya2024novel, zhang2024comparative, kilincer2023automated, gupta2022tree,  fouda2022novel, alsalman2024comparative, saheed2021efficient, manimurugan2020effective, hady2020intrusion, ravi2023deep, khan2023secure, nandy2021intrusion, gao2017machine, he2019intrusion, newaz2020heka,  areia2024iomt, ioannou2024gemlids, sohail2024explainable}, (ii) evaluation with non IoMT or simulated datasets \cite{rm2020effective, alrashdi2019fbad, kumar2021ensemble, khan2021hybrid, khan2022xsru, gao2017machine, he2019intrusion, newaz2020heka, begli2019layered}, (iii) coarse-grained classification that cannot distinguish between attacks types \cite{wang2024federated, dhanya2024novel, zhang2024comparative, kumar2021ensemble, rm2020effective, kilincer2023automated, gupta2022tree, fouda2022novel, alsalman2024comparative, saheed2021efficient, hady2020intrusion, ravi2023deep, khan2023secure, nandy2021intrusion}, and (iv) insufficient classification granularity or even lack of zero-day attack detection and classification \cite{rm2020effective, alrashdi2019fbad, kumar2021ensemble, khan2021hybrid, khan2022xsru, wang2024federated, dhanya2024novel, zhang2024comparative, kilincer2023automated, gupta2022tree,  fouda2022novel, alsalman2024comparative, saheed2021efficient, manimurugan2020effective, hady2020intrusion, ravi2023deep, khan2023secure, nandy2021intrusion, gao2017machine, he2019intrusion, newaz2020heka,  areia2024iomt, ioannou2024gemlids, sohail2024explainable, begli2019layered, thamilarasu2020intrusion, gupta2022cybersecurity, singh2022dew, zukaib2024meta, dadkhah2024ciciomt2024, ramesh2024efficient, mohammadi2024securing, vu2024performance}. 
    These limitations create significant security gaps in IoMT environments, necessitating a new approach tailored to medical IoT constraints

To address these gaps, we propose a novel hierarchical intrusion detection framework that accurately detects both known and zero-day attacks by strategically distributing specialized classifiers across different IoMT network layers. 
    This hierarchical approach can reduce computational overhead on resource-constrained medical devices while maintaining high detection accuracy and enabling real-time threat response
    In the next section, we will discuss our proposed IDS in detail.

\section{The Proposed Hierarchical IDS Model}
\label{Proposed}

\subsection{Proposed IDS Model for Zero-day Attack Detection}
Our analysis of the current state-of-the-art research reveals that most studies on IDS in IoMT rely on conventional flat and centralized multi-class classification approaches. 
    As such, these models may suffer from bottleneck issues at central processing points, and they does not scale effectively across distributed IoMT networks.
    Furthermore, supervised models are built on historical data, and thus this limits their effectiveness in detecting zero-day attacks because the model can only learn the behaviors of normal and historical attacks. 
    In fact, zero-day attacks (which are increasingly prevalent in modern networks) often differ from the historical attacks.

\begin{figure}[!htbp]
   \centering
    \includegraphics[width = \linewidth]{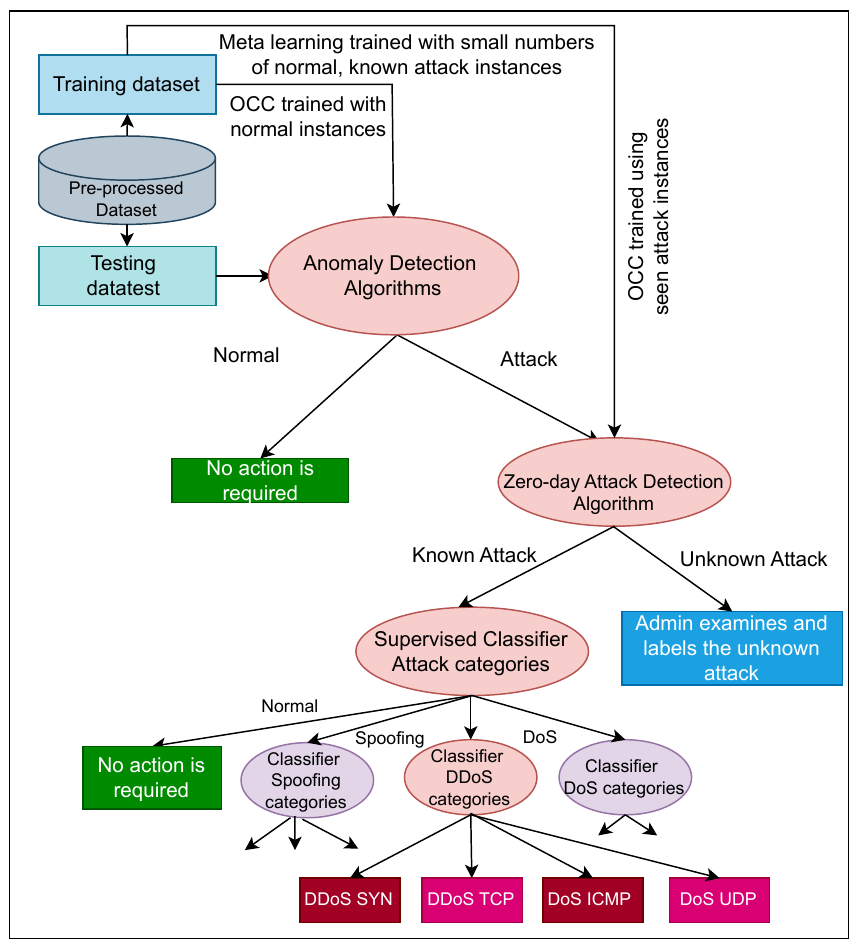}
    \caption{The proposed zero-day detection enabled IDS framework.}
    \label{fig:metamodel}
\end{figure}

To address these challenges, we recommend a hierarchical intrusion detection system designed to detect cyberattacks at multiple levels in IoMT using meta learning or OCC and conventional supervised classifiers. 
    Figure \ref{fig:metamodel} illustrates the proposed framework, which begins with a pre-processed dataset serving as input for training and testing. 
    At the root level of the hierarchy, we suggest to employ meta learning or OCC to detect anomaly. 
    Here, meta learning leverages very minimal training data for detecting anomaly effectively. 
    On the other hand, if we use OCC such as usfAD or other semi-supervised models, we only require benign network traffic to train the model which can address the class imbalanced issue. {Lightweight models are recommended at the root level for deployment on Near Edge nodes due to their computational efficiency and adaptability are critical.}

To train the meta learning-based classifier at root level, we divide the dataset into smaller datasets, each consisting of normal traffic and one attack type (e.g., DDoS). 
    Then, we create multiple similar classification tasks in which each task is to recognize one attack type. 
    For example, $Task_1$ is to classify the traffic as DDoS or Normal, $Task_2$ is to classify the traffic as Spoofing or Normal, and so on. 
    Here, we apply a lightweight meta-learning algorithm called Reptile~\cite{nichol2018first} (which is presented in Algorithm~\ref{alg:meta_learning}) to train the root classifier with these tasks.
    By doing so, the root classifier can generalize  across different attack patterns and thus effectively detect zero-day/unknown attacks. 
    Note that our proposed meta-learning classifier only requires less than 1\% of the CICIoMT2024 dataset while still achieving high detection performance, thus significantly reducing the computing and storage overhead.
    Once deployed, the meta learning classifier determines whether an input is normal or indicative of an attack. 
    The key advantage of meta learning is its ability to detect zero-day/unknown attacks without requiring retraining or new datasets.
    
As shown in Fig.~\ref{fig:metamodel}, at the root level, if a network traffic is classified as normal, no further action is taken. 
    If a network traffic is classified as an anomaly/attack, it is passed to the next layer of the IDS for further analysis to determine if it is a known or unknown attack. 
    In this work, we also suggest to use a specific model called usfAD at the root level to detect anomaly. 
    The advantage of using this model is that there needs only benign or normal network traffic to train it. 
    Our experiment shows that usfAD at the root level achieves higher accuracy than other OCC models and meta learning.  
    At the second level/layer, for detecting known and unknown attack types, we propose training a one-class classifier. 
    This layer distinguishes known attacks and directs them to supervised classifiers for detailed categorization in the Far Edge nodes. 
    Then, unknown attacks are collected and analyzed by the security analyst. 
    Unknown attacks are labeled based on their characteristics and used to retrain the supervised model.  
    In the Cloud (i.e., the third layer), multiple supervised classifier are used to identify subcategories of known attack types (e.g., Spoofing, DoS, DDoS). 
    Once the specific attack type is identified, a reporting mechanism is triggered to log the incident and initiate appropriate responses. 
    Given the above, our proposed hierarchical IDS can effectively capture zero-day attacks, provide insights about attack patterns, and update its knowledge accordingly.

\begin{algorithm}[t]
\caption{Meta learning-based Classifier}
\label{alg:meta_learning}
\begin{algorithmic}[1]
\State Initialize $\Theta$, the initial parameters of a Deep Neural Network (DNN)
\For{iteration $i = 1, 2, 3, \ldots$}
    \State Randomly sample a task $T$ from the set of anomally detection tasks $\mathcal{T}_i = \{T_1, T_2, \ldots\}$
    \State Perform $k$ steps of Stochastic Gradient Descent (SGD) on task $T_i$, starting with parameters $\Theta_i$, resulting in parameters $\Omega_i$
    \State Update: $\Theta_{i+1} \leftarrow \Theta_i + \epsilon(\Omega_i - \Theta_i)$ \Comment{$\epsilon$ is the meta-learning rate}
\EndFor
\State \Return{$\Theta_i$}
\end{algorithmic}
\end{algorithm}

Our hierarchical design combines the strengths of meta-learning or OCC at the root level for general detection with supervised classifiers for granular attack classification in the subsequent layers. This approach ensures efficient and accurate detection, especially in resource-constrained environments like medical end devices, while addressing the critical challenge of detecting zero-day attacks in distributed networks.
    The details of our proposed approach are presented in Algorithm~\ref{alg:hids}


The data processing steps are briefly described below. 

\begin{itemize}
    \item Dataset: We utilized the CICIoMT2024 dataset \cite{dadkhah2024ciciomt2024}, a comprehensive multi-protocol dataset, to train and test a hierarchical IDS for IoMT devices. 
        This dataset, sourced from the Canadian Institute for Cybersecurity, contains data from both Wi-Fi-enabled IoMT devices and simulated MQTT-based devices.
        In order to  form the dataset, a range of cyberattacks was executed against IoMT devices. 
        The dataset provider encompassed five main attack categories named as DDoS, DoS, Reconnaissance, Spoofing, and MQTT-based attacks. Each category is further divided into 18 specific attack subtypes, offering a detailed and diverse dataset for analysis. To ensure the dataset's suitability for our experiments, we meticulously preprocessed the provided training and testing data, aligning it with the requirements of our intrusion detection framework.

    Table \ref{tab:dataset} shows the distribution of classes in the dataset \cite{dadkhah2024ciciomt2024}. 11.19\% of the records are normal, while the remaining records represent various types of attacks. Ping Sweep records have the smallest contribution, accounting for less than 0.01\% of the total records. The highest percentage of records belong to DDoS ICMP class (20.12\%).

    \item Data Scaling: We apply Min-Max normalization that eliminates the impact of different feature's value in the datasets on the performance of machine learning algorithms. 
        Min-Max approach scales each feature's values to a range between 0 and 1. 
        The min-max normalisation formula is follows: $X_{norm} = \frac{X - X_{min}}{X_{max} - X_{min}}$, where $X_{norm}$, $X_{min}$, $X_{max}$  are the normalized value, the minimum value, and the maximum value of $X$, respectively.
   
   \item Dataset Partition: Stratified cross-validation is used to divide the dataset into folds while maintaining the proportional distribution of each class within each fold. 
        This method provides a reliable estimate of model performance, especially when dealing with imbalanced datasets where one class has significantly more samples than others \cite{kohavi1995study}. 
        In this study, our hierarchical model is trained and evaluated using stratified 10-fold cross-validation. 
        Specifically, the dataset was split into ten folds of equal size, with each fold preserving the class proportions. 
        This process was repeated for all ten folds, and the average classification performance was reported. 
        By utilizing stratified cross-validation, both models (meta learning and OCC) achieve more accurate and dependable performance evaluations. 
\end{itemize}

\begin{table}[t]
\caption{Class distributions in the dataset}
\label{tab:dataset}
\centering

\begin{tabular}{|l | l | r | r |}
\hline
\multicolumn{2}{|l|}{Class} & Records & Percentage \\
\hline
\multicolumn{2}{|l|}{Benign} & 1048575 & 11.19 \\
\hline
Spoofing	&ARP Spoofing & 717791	&0.19\\
\hline

\multirow{4}{*}{\begin{sideways}Recon \end{sideways}}& Ping Sweep	& 926	& 0.01 \\
\cline{2-4}
&Recon VulScan &	3207 &	0.03 \\
\cline{2-4}
&OS Scan	& 20666	& 0.22 \\
\cline{2-4}
&Port Scan	& 106603	& 1.14 \\
\hline
\multirow{4}{*}{\begin{sideways}MQTT \end{sideways}}& Malformed Data &	6877 &	0.07 \\
\cline{2-4}
&DoS Connect Flood	&15904	& 0.17 \\
\cline{2-4}
&DDoS Publish Flood	& 36039 &	0.38 \\
\cline{2-4}
& DoS Publish Flood	& 52881	& 0.56 \\
\hline
\multirow{4}{*}{\begin{sideways}DoS \end{sideways}}& DoS TCP	& 462480	& 4.94 \\
\cline{2-4}
&DoS ICMP	& 514724	& 5.49 \\
\cline{2-4}
&DoS SYN	& 540498	& 5.76 \\
\cline{2-4}
& DoS UDP	& 704503	& 7.51 \\
\hline
\multirow{4}{*}{\begin{sideways}DDoS \end{sideways}}& DDoS SYN	& 974359	& 10.39 \\
\cline{2-4}
&DDoS TCP	& 987063	& 10.53 \\
\cline{2-4}
&DoS ICMP	& 514724	& 5.49 \\
\cline{2-4}
& DoS UDP	& 1998026	& 21.3 \\
\hline
\multicolumn{2}{|l|}{Total} & 9378297
  & 100 \\
\hline
\end{tabular}
\end{table}






\begin{algorithm}[t]
\caption{Hierarchical Intrusion Detection System (H-IDS) for Detecting Known and Zero-Day Attacks}
\label{alg:hids}
\begin{algorithmic}[1]
\Require Preprocessed dataset $D$
\Ensure Classification of each instance $x \in D$ as \texttt{Normal}, \texttt{Known Attack}, or \texttt{Unknown Attack}

\State \textbf{Training Phase:}
\State Train $M_{\text{root}}$ (e.g., usfAD) on normal instances from $D$ \Comment{Root-level anomaly detector}
\State Train $M_{\text{verify}}$ using known attack instances \Comment{Verifies known vs unknown attack}
\State Train supervised classifiers for each attack category:
\hspace{\algorithmicindent} $M_{\text{DoS}}$, $M_{\text{DDoS}}$, $M_{\text{Spoof}}$, etc.
\State
\State \textbf{Detection Phase (for each incoming instance $x$):}
 $result_{\text{root}} \leftarrow M_{\text{root}}(x)$
\If{$result_{\text{root}} ==$ Normal}
    \State \Return \texttt{Normal} \Comment{No further action needed}
\Else
    \State $result_{\text{verify}} \leftarrow M_{\text{verify}}(x)$
    \If{$result_{\text{verify}} ==$ Known Attack}
        \State $type \leftarrow$ detect attack category of $x$
        \State $subtype \leftarrow M_{type}(x)$ \Comment{Classify into subtype}
        \State Log incident and trigger response
        \State \Return $subtype$
    \Else
        \State Forward $x$ to security analyst for manual inspection
        \State Label $x$ as \texttt{Unknown Attack} and store for future retraining
        \State \Return \texttt{Unknown Attack}
    \EndIf
\EndIf
\end{algorithmic}
\end{algorithm}

\subsection{Distributing the Proposed Hierarchical IDS in IoMT}

\begin{figure}[t]
   \centering
    \includegraphics[width = \linewidth]{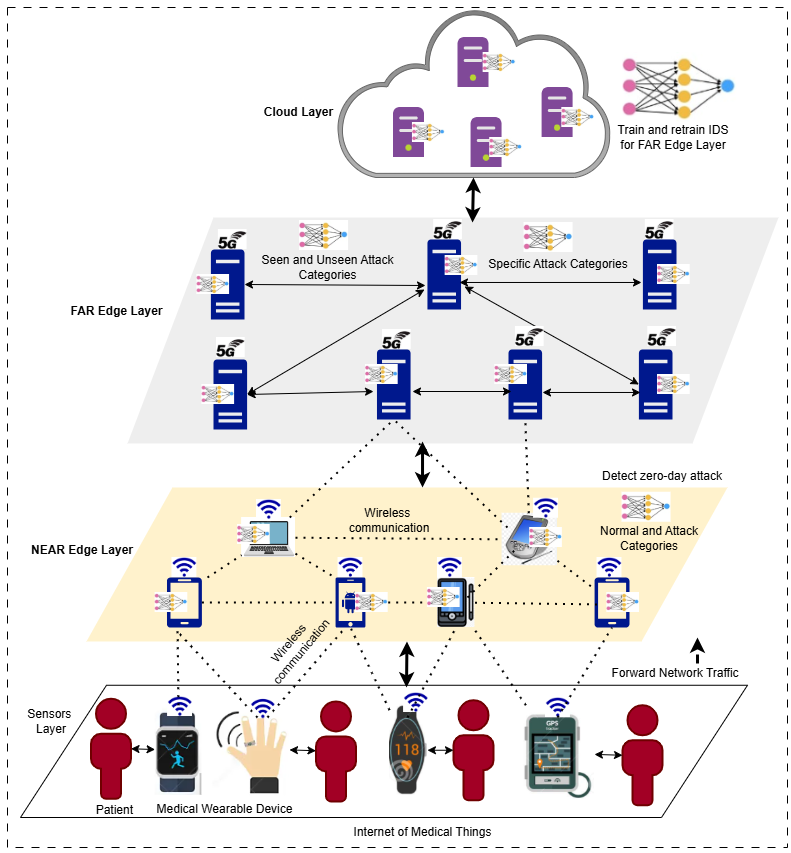}
    \caption{The proposed architecture of IDS enabled Internet of Medical Things}
    \label{fig:iomtarchitecture}
\end{figure}

The IoMT is continually exposed to emerging threats, including zero-day attacks, which places patients' data at constant risk of privacy and security breaches. 
    A main reason is that the IoMT comprises heterogeneous devices, e.g, wearable sensors, medical end devices (such as mobile phones and PDAs), Edge nodes, and Cloud servers. 
    The typical IoMT network presented in Fig. \ref{fig:iomtarchitecture} consists of different layers or levels called Sensors, Near Edge, Far Edge and Cloud layers to provide real-time health monitoring and data analysis in a distributed and resource-efficient manner \cite{rahmani2018exploiting}. 
    Medical devices transmit health data to the Cloud via the Near and Far Edge layers, while also receiving network traffic from upper layers through the Near Edge nodes.
    To protect patient data, an IDS should be deployed in medical devices. 
    However, medical devices are typically resource-constrained and cannot support such systems locally. 
    The nearest devices to these medical devices (e.g., smartphones, PDAs, and user computers that collectively referred to as the Near Edge) offer a feasible alternative. 
    As such, deploying IDS at the Near Edge enables rapid detection of potential attacks and helps safeguard the medical devices. 
    However, leveraging conventional supervised learning models (e.g., Random Forest) is not a viable solution as they may be ineffective at detecting zero-day attacks. 
    Moreover, these supervised models require large volumes of labeled benign and attack data for training, which Near Edge devices typically do not possess. 
    Therefore, it is very important to effectively detect known and zero-day attacks at NEAR edge devices to protect both patients and their data. 

To address this issue, we devise a hierarchical IDS approach that aligns well with the IoMT's hierarchical architecture, allowing different components to be strategically deployed across various layers. 
    By doing so, our approach can optimize computational resources and enhance scalability in IoMT \cite{rahmani2015smart}. 
    Figure \ref{fig:iomtarchitecture} illustrates our proposed hierarchical architecture for an IDS designed for the Internet of Medical Things (IoMT).     
    It showcases how IDS components are distributed strategically across the different layers of the IoMT infrastructure to ensure robust security while maintaining efficiency.

\begin{itemize}
   \item \textbf{Sensors Layer}: This is the bottom-most layer, consisting of patients, medical wearable devices, and IoMT sensors. 
        Devices such as fitness trackers, heart rate monitors, and GPS-enabled devices collect real-time data from patients. 
        These devices forward network traffic to the upper layers for further analysis and security checks.
    
    \item \textbf{Near Edge Layer:} This layer is positioned above the Sensors layer.
        In our proposed hierarchical IDS, the trained root classifier (based on meta learning or OCC) of the hierarchical framework is deployed at the Near Edge layer, within medical end devices such as smartphones or local IoMT hubs. 
        This root classifier distinguishes between normal and malicious data (i.e., anomaly), enabling immediate detection and response to cyber threats. 
        By processing data locally, this setup reduces delays caused by communication overhead and ensures rapid decision-making, reducing latency and dependency on centralized servers. 
        Note that the root classifier does not determine whether the detected malicious or anomalous instances correspond to known attacks or previously unknown threats (e.g., zero-day attacks).
     
    \item \textbf{Far Edge Layer:} Far Edge layer acts as an intermediary layer, connecting the NEAR Edge Layer with the Cloud Layer. 
        FAR Edge devices are high-performance nodes (e.g., 5G servers) compared to Near Edge nodes and capable of handling more computationally intensive tasks. 
         Given that, we suggest the Far Edge layer for deploying the second-level classification model. 
        Specifically, the Far Edge layer aggregates and processes data from multiple medical devices, enabling the classification of threats into broader categories such as zero-day and known attacks.
        In this work, a One-Class Classification (OCC) model is leveraged to distinguish between known and unknown (i.e., zero-day) attacks.     
        If an attack is identified as known, a subsequent supervised classifier at the Cloud layer further categorizes the specific type of attack. 
        This intermediate layer significantly enhances the system’s ability to manage diverse data sources while ensuring timely and accurate threat analysis.
        
    \item \textbf{Cloud Layer:} The top-most layer is the Cloud that provides the computational resources needed for tasks requiring high processing power, such as model training and large-scale data analysis. 
    As such, we recommend Cloud layer for deploying the top-level classification models that detect detailed attack subcategories and serves as the central hub for training and retraining the hierarchical classification model. 
    This top level ensures that intrusion detection models deployed across the IoMT network remain up-to-date and capable of addressing evolving threats.
    This top level ensures that intrusion detection models deployed across the IoMT network remain up-to-date and capable of addressing evolving threats.
    
\end{itemize}

Unlike traditional centralized flat classification models (e.g., \cite{rm2020effective, alrashdi2019fbad, kumar2021ensemble, khan2021hybrid, khan2022xsru, wang2024federated, dhanya2024novel, zhang2024comparative, kilincer2023automated, gupta2022tree,  fouda2022novel, alsalman2024comparative, saheed2021efficient, manimurugan2020effective, hady2020intrusion, ravi2023deep, khan2023secure, nandy2021intrusion, gao2017machine, he2019intrusion, newaz2020heka,  areia2024iomt, ioannou2024gemlids, sohail2024explainable}), which rely on a single point of processing and cannot be distributed across IoMT network layers, our proposed hierarchical model leverages the IoMT's layered architecture to distribute tasks strategically and efficiently. 
    Although hierarchical models can involve additional classifiers and higher complexity, they offer significant advantages (including enhanced scalability, reduced communication overhead, and improved adaptability) that can outweigh these limitations in distributed IoMT environments. 
    This hierarchical architecture enables strategic distribution of system components across network tiers, ensuring robust intrusion detection while maintaining the lightweight functionality required for medical devices operating under resource constraints.



\section{Results and Discussion}
\label{RESULTS AND DISCUSSION}

This section will evaluate the effectiveness of the hierarchical model for IoMT in terms of accuracy, precision, recall and F1-score. 
    We target to optimize resource usage by delegating tasks to the most appropriate layer of IoMT based on computational requirements and urgency. 
    In fact, this architecture can be easily adapted to handle more devices and attack scenarios by distributing the workload across layers. 
    By detecting zero-day attacks at the NEAR Edge layer and refining classifications at higher layers, the system ensures quick and accurate responses.

\subsection{Experimental Setup and Implementation}

Our experimental study was performed on an Intel Xeon E5-2670 CPU (8 cores, 16 threads), 128GB DDR3 RAM, 2x Nvidia GTX 1080 Ti. 
    Python 3.9 was used to execute our code. 
    This study utilized meta learning, OCC based usfAD~\cite{aryal2021usfad}, Random Forest and primarily relied on Pandas and NumPy libraries for data pre-processing. 
    Since the framework was developed using Python, the widely recognized Scikit-learn toolkit was utilized to leverage its wide range of algorithms and resources for data scientists, including effective accuracy and precision estimation metrics. 
    In this work, we employed machine-learning algorithms from Scikit-learn and usfAD algorithm\cite{aryal2021usfad}.

\subsection{Performance Metrics}

In this study, we used accuracy, precision, recall, and F1-score that are essential for assessing the performance of an IDS model. 
    However, their significance can vary depending on the system's specific objectives and requirements. 
    Specifically, accuracy quantifies the proportion of accurate classifications made by the IDS. 
    However, relying solely on accuracy is not the most suitable performance metric for IDS as this might not accurately reflect the system's capability to identify attacks that belong to a minority class within the dataset. 
    On the other hand, precision refers to the proportion of genuine positive detection out of all positive detection. 
    High precision is essential in IDS in order to minimize false negatives (normal transactions are incorrectly detected as attack), which can result in false alarms. 
    Differently. recall measures the system's ability to reliably identify all instances of a particular class of attack. 
    Low recall suggests that the system is missing many attacks, which can pose a significant security risk.
    The F1-score is a combination of precision and recall that quantifies the proportion of true positive identification relative to the total number of positive instances in the dataset. 
    F1-score is a valuable metric for IDS because this considers both false positives and false negatives and provides a balanced score between precision and recall.

The accuracy, precision, recall and F1-score are calculated as follows.

\[
\text{Accuracy} = \frac{\text{TP + TN}}{\text{TP + TN + FP + FN}} \times 100,
\]

\[
\text{Precision} = \frac{\text{TP}}{\text{TP+FP}} \times 100,
\]

\[
\text{Recall} = \frac{\text{TP}}{\text{TP+FN}} \times 100,
\]

\[
\text{F1-score} = 2\times \frac{\text{Precision $\times$ Recall}}{\text{Precision + Recall}} \times 100,
\]

where TP = true positive, TN = true negative, FP = false positive, and FN = false negative.

In our experiment, we used Stratified cross-validation which can effectively address the imbalance in test datasets by maintaining balanced class distributions across folds, thus ensuring accurate evaluation metrics such as precision, recall, and F1-score. 
    This approach guards against biased evaluations driven by dominant classes and enhances model robustness to dataset variability. 
    Overall, it provides reliable estimates of generalization performance for models trained on imbalanced data, crucial for real-world applications. 
    In the subsequent sections, we present the performance of our hierarchical IDS model in IoMT.

\subsection{Root Level Classifier's Performance to Detect Attacks}

In this section, we compare the performance of meta-learning and several OCC algorithms in detecting normal and attack instances.    
    For the meta-learning model, we use both normal and different attack categories from the dataset for training the model.
    Our approach involves creating multiple sub-datasets for training the root classifier, where each sub-dataset consists of 128 instances from one attack type and 128 instances from benign data. 
    We then define multiple similar classification tasks for meta-learning, with each task designed to recognize one specific attack type.    
    For example, $Task_1$ is to classify the traffic as DDoS or Normal, $Task_2$ is to classify the traffic as Spoofing or Normal, and so on. 
    Here, we adopt a lightweight meta-learning algorithm, namely Reptile~\cite{nichol2018first}, to train the root classifier across these tasks, enabling it to generalize across different attack patterns and effectively detect zero-day/unknown attacks.
    The hyperparameters are similar to those in~\cite{nichol2018first}, e.g., $k = 5$. 
    Our framework leverages a simple, small-footprint DNN model with only three layers, with the hidden layer consisting of 64 neurons.
    To simulate zero-day attack scenarios, we train and test the models multiple times. 
    Each time, we systematically exclude one attack type from the training process, treating it as an unknown attack. 
    For instance, in one scenario, we train the root classifier with all tasks except DoS, making DoS completely unknown to the model during testing. 
   To evaluate the effectiveness of the meta-learning approach, conventional Stochastic Gradient Descent (SGD), a well-known algorithm for training DNNs~\cite{goodfellow2016deep}, is also employed under the same experimental setup.
   By doing so, this experimental setup demonstrates the meta-learning algorithm's capability to detect anomalies, including zero-day/unknown attacks.

 We used the following models for the OCC, all of which were trained using only normal instances. 
 
   \begin{itemize}
       \item Unsupervised Stochastic Forest-based Anomaly Detector (usfAD): This is a robust anomaly detection algorithm based on an unsupervised stochastic forest. Aryal et al. \cite{aryal2021usfad} demonstrated the effectiveness of this approach in detecting anomalies across various cybersecurity datasets.
       \item Local Outlier Factor (LOF): This algorithm assigns a degree of outlier-ness to each data object, referred to as the local outlier factor (LOF). The LOF is local in nature and reflects the level of isolation of a data object within its neighbourhood. Breunig et al.  \cite{breunig2000lof} demonstrated the success of this algorithm in identifying local outliers.
       \item Isolation Forest (IF): While many model-based anomaly detection algorithms rely on building a profile of normal instances, the Isolation Forest (iForest) isolates anomalies without requiring such a profile. iForest makes effective use of sub-sampling and has been shown by Liu et al. \cite{liu2008isolation} to perform well on high-dimensional datasets with many irrelevant features, as well as on datasets lacking labelled anomaly instances.
   \end{itemize}


The results in Table \ref{tab:rootlevel} present the performances of the root classifier with different algorithms in detecting normal and attack at the root level. 
    The usfAD outperforms the other OCC and meta-learning models with a higher accuracy (i.e., 99.77\%) and near-perfect precision, recall, and F1-scores for both normal and attack classes. 
    This demonstrates a well-balanced and robust performance of the OCC based usfAD. 
    Additionally, its high Macro and Weighted Average F1-scores further confirm its reliability for scenarios requiring accurate detection of both normal and attack instances.

\begin{table*}[t]
\caption{Performance comparison of One Class classifiers and Meta Learning to detect attacks}
    \centering
    \begin{tabular}{|c|c|l|l|l|l|}
        \hline
        \textbf{Root Classifier} & \textbf{Accuracy} & \textbf{Detail} & \textbf{Precision} & \textbf{Recall} & \textbf{F1-score} \\ \hline
        \multirow{4}{*}{usfAD} & \multirow{4}{*}{99.77} & Normal & 93.4 & 98.2 & 95.74 \\ \cline{3-6}
        &  & Attack & 99.95 & 99.81 & 99.88 \\ \cline{3-6}
        &  & Macro Avg & 96.68 & 99 & 97.81 \\ \cline{3-6}
        &  & Weighted Avg & 99.78 & 99.77 & 99.77 \\ \hline
        
        \multirow{4}{*}{Local Outlier Factor (LOF)} & \multirow{4}{*}{99.46} & Normal & 88.24 & 91.5 & 89.84 \\ \cline{3-6}
        &  & Attack & 99.77 & 99.67 & 99.72 \\ \cline{3-6}
        &  & Macro Avg & 94 & 95.58 & 94.78 \\ \cline{3-6}
        &  & Weighted Avg & 99.47 & 99.46 & 99.46 \\ \hline
        
        \multirow{4}{*}{Isolation Forest (IF)} & \multirow{4}{*}{82.34} & Normal & 11.51 & 85.65 & 20.3 \\ \cline{3-6}
        &  & Attack & 99.53 & 82.55 & 90.55 \\ \cline{3-6}
        &  & Macro Avg & 55.52 & 83.95 & 55.18 \\ \cline{3-6}
        &  & Weighted Avg & 97.22 & 82.34 & 90.74 \\ \hline
        
        \multirow{4}{*}{Meta-learning Classifier (MLC)}  & \multirow{4}{*}{98.27} & Normal & 98.01 & 95.52 & 96.69 \\ \cline{3-6}
        &  & Attack & 97.65 & 97.56 & 97.51 \\ \cline{3-6}
        &  & Macro Avg & 97.83 & 96.54 & 97.10 \\ \cline{3-6}
        &  & Weighted Avg & 97.66 & 97.51 & 97.50 \\ \hline

        \multirow{4}{*}{Stochastic Gradient Descent (SGD)}  & \multirow{4}{*}{1.241} & Normal & 27.44 & 5.44 & 6 \\ \cline{3-6}
        &  & Attack & 0.33 & 0.32 & 0.065 \\ \cline{3-6}
        &  & Macro Avg & 13.89 & 2.88 & 3.03 \\ \cline{3-6}
        &  & Weighted Avg & 1.05 & 0.45 & 0.22 \\ \hline
        
    \end{tabular}
    
    \label{tab:rootlevel}
\end{table*}


\begin{figure}[t]
   \centering
    \includegraphics[width = .95\linewidth]{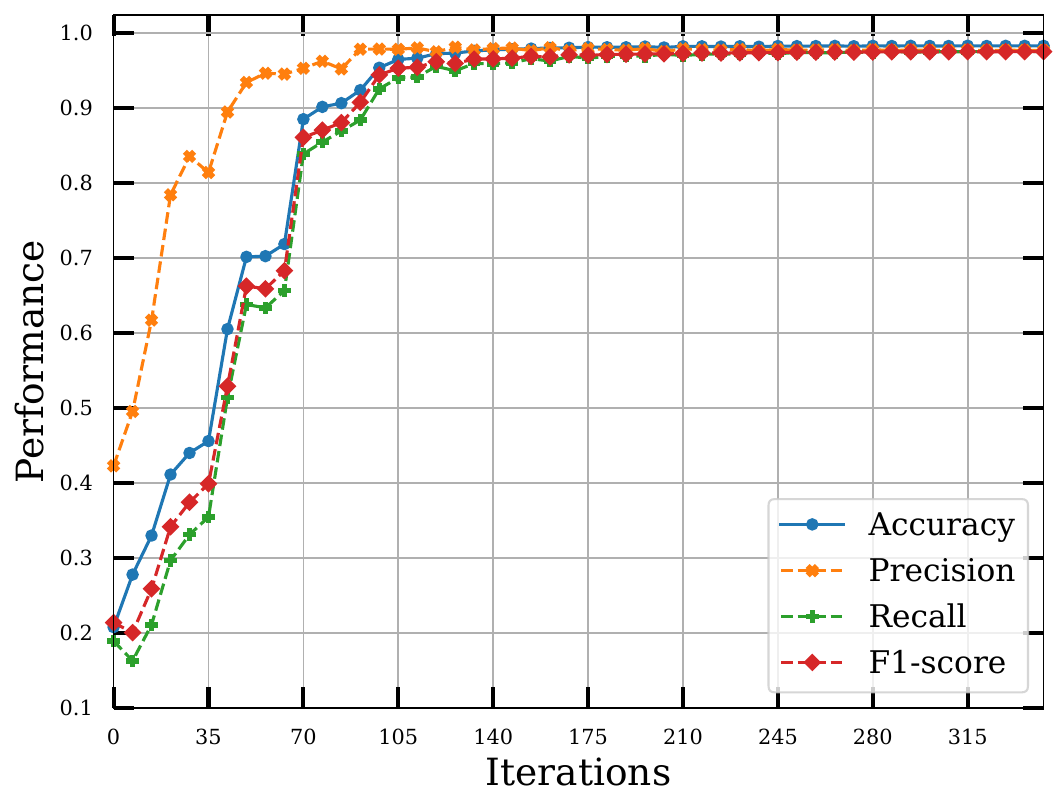}
    \caption{The learning process of meta-learning}
    \label{fig:meta-learning-curve}
\end{figure} 

The second OCC model, i.e., Local Outlier Factor (LOF), also performs well with high accuracy of 99.46\% and strong attack detection metrics but slightly struggles in identifying normal instances, as reflected by its lower precision and recall for the normal class compared with those of usfAD. 
    However, the third OCC model, i.e., Isolation Forest (IF) exhibits significant weaknesses, with poor detection of normal instances (F1-score: 20.3\%) and an overall accuracy of only 82.34\%, making it unsuitable for imbalanced detection tasks.

Interestingly, thanks to its "learn to learn" ability, the meta-learning model achieves good accuracy (98.27\%) using less than 1\% of the total dataset. 
    On the other hand, it clearly shows that the traditional deep learning algorithm (i.e., SGD) is unable to detect anomalies as well as zero-day attacks when trained with a small dataset.
    Additionally, Fig.~\ref{fig:meta-learning-curve} shows that MLC quickly achieves a high overall performance after 175 iterations.
    While MLC's accuracy is lower than those of usfAD and LOF, it significantly outperforms IF. 
    Notably, meta-learning surpasses all OCC models in detecting normal instances, clearly demonstrating its potential for zero-day attack detection.
    
Given these results, usfAD emerges as the most effective and reliable classifier for detecting attack instances. 
    On the other hand, the meta-learning algorithm offers distinct advantages that make it more suitable for resource-constrained environments: lower training complexity and minimal data requirements. 
    This efficiency is particularly valuable for Near Edge devices, where large datasets may not be available for retraining. 
    By deploying meta-learning on such devices, we can effectively detect unknown attacks while addressing the time and resource constraints of low-profile devices.



\subsection{OCC's Performance for Detecting Unknown(Zero-day) and Known Attack}


Although we recommend that the meta-learning algorithm can be deployed on Near Edge devices to detect both historical (known) and zero-day (unknown) attacks, it faces a fundamental limitation in distinguishing between known and unknown attacks, a distinction that is critical for model retraining and threat intelligence.
    This limitation arises from the binary classification nature of our meta-learning approach, which is designed to distinguish only between normal traffic and attack instances, without further categorizing attacks as known or unknown.
    More critically, training the meta-learning model with actual zero-day attack data would violate the fundamental principle of zero-day attacks—that they are previously unseen and unavailable during training.

Given the resource constraints of Near Edge devices, a simple binary classifier is appropriate for this layer.
    In this context, the OCC is a promising solution.
    Since OCC is trained exclusively on known attack instances, it can identify anomalous or previously unseen (unknown) attacks effectively.
    Therefore, this work leverage the OCC in the Far Edge layer, where it can distinguish unknown attacks from historical ones, thereby supporting targeted model retraining and further improving detection accuracy.
   
Table \ref{tab:secondlevel} presents the performance of the one-class classifiers (including usfAD, LOF, and IF) in detecting known and unknown attack types, focusing on detecting zero-day attacks as the second-level detection in the framework. 
    In this experiment, we trained and tested the models multiple times. 
    Each time, we assumed a particular attack unknown and all other attacks known. For example, we make DoS attack unknown by removing the instances of DoS from the training set but retraining this attack and other attack instances in the testing set. 
    Then, we repeat this process with other attack types, i.e., MQTT, DDoS, Recon, and Spoofing. 
    Across the five attack categories, usfAD demonstrates superior overall performance in both known and unknown (zero-day) attack detection, achieving high precision, recall, and F1-scores, 
    particularly excelling in detecting unknown attacks. 
    For instance, usfAD achieves an F1-score of 91.03\% for DoS attacks when it is unknown and 88.48\% for unknown MQTT attacks, significantly outperforming LOF and IF in these scenarios.

LOF, on the other hand, shows limited effectiveness for unknown attacks, with extremely low F1-scores (e.g., 15.2\% for DoS and 2.81\% for MQTT), although its performance is good for known attacks (e.g., an F1-score of 84.48\% for MQTT).
    This suggests that LOF struggles with identifying patterns in unseen data and might be more suited for datasets dominated by known attack instances. 
    In contrast, IF exhibits mixed performance, excelling in specific unknown attack scenarios (e.g., an F1-score of 87.5\% for unknown MQTT) but failing significantly in others (e.g., an F1-score of 0.76\% for unknown DDoS and 23.18\% for unknown Spoofing). 
    This variability indicates a lack of robustness across different attack types.

Overall, usfAD emerges as the most reliable classifier for detecting both known and unknown attacks, making it a strong candidate for the second-level zero-day detection in the framework. Its ability to balance precision and recall across diverse attack types ensures robust performance. In contrast, LOF and IF show inconsistent results, with limited applicability for detecting unknown attacks, which underscores the importance of selecting classifiers tailored to the detection task at hand.

\begin{table*}[t]
    \centering
    \caption{One Class Classifiers' performance to detect known and unknown attack types}
    \label{tab:secondlevel}
    \renewcommand{\arraystretch}{1.2} 
    \resizebox{\textwidth}{!}{ 
    \begin{tabular}{|p{2.1cm}|p{1.5cm}|p{1.2cm}|p{1.2cm}|p{1.2cm}|p{1.2cm}|p{1.2cm}|p{1.2cm}|p{1.2cm}|p{1.2cm}|p{1.2cm}|}
        \hline
        \multirow{2}{*}{\textbf{Attack Type}} & 
        \multirow{2}{*}{\textbf{Class}} & 
        \multicolumn{3}{c|}{\textbf{usfAD}} & 
        \multicolumn{3}{c|}{\textbf{LOF}} & 
        \multicolumn{3}{c|}{\textbf{IF}} \\
        \cline{3-11}
         &  & \textbf{Prec.} & \textbf{Recall} & \textbf{F1} & 
         \textbf{Prec.} & \textbf{Recall} & \textbf{F1} & 
         \textbf{Prec.} & \textbf{Recall} & \textbf{F1} \\
        \hline
        \multirow{2}{*}{DoS} 
            & Unknown & 96.17 & 86.42 & 91.03 & 54.39 & 8.84 & 15.20 & 15.25 & 1.02 & 1.92 \\
            & Known   & 84.52 & 95.57 & 89.70 & 43.50 & 90.45 & 58.74 & 42.08 & 92.68 & 57.88 \\
        \hline
        \multirow{2}{*}{MQTT} 
            & Unknown & 82.42 & 95.49 & 88.48 & 4.23  & 2.10  & 2.81  & 78.55 & 98.76 & 87.50 \\
            & Known   & 98.95 & 95.41 & 97.15 & 80.18 & 89.26 & 84.48 & 99.70 & 93.92 & 96.73 \\
        \hline
        \multirow{2}{*}{DDoS} 
            & Unknown & 72.04 & 52.84 & 60.96 & 18.49 & 10.01 & 12.99 & 1.56  & 0.50  & 0.76 \\
            & Known   & 90.91 & 95.83 & 93.30 & 83.27 & 91.03 & 86.98 & 82.22 & 93.54 & 87.51 \\
        \hline
        \multirow{2}{*}{Recon} 
            & Unknown & 69.44 & 98.64 & 81.51 & 21.71 & 26.84 & 24.01 & 42.63 & 57.11 & 48.82 \\
            & Known   & 99.89 & 96.53 & 98.18 & 94.04 & 92.27 & 93.15 & 96.48 & 93.86 & 95.15 \\
        \hline
        \multirow{2}{*}{Spoofing} 
            & Unknown & 16.13 & 70.97 & 26.29 & 1.18  & 10.75 & 2.12  & 13.20 & 94.82 & 23.18 \\
            & Known   & 99.70 & 96.27 & 97.95 & 99.02 & 90.87 & 94.77 & 99.94 & 93.69 & 96.72 \\
        \hline
    \end{tabular}
    }
\end{table*}


Unlike one-class classifiers, supervised classifiers like Random Forest (RF) fail to detect unknown attacks because they require training on specific types of instances to recognize those patterns~\cite{uddin2024usfad}. 
    This means that the RF model, as a supervised learning algorithm, is limited to the patterns seen in its training data and struggles to generalize beyond them. 
    As a result, it is not effective at identifying zero-day or unknown attacks.

\begin{figure}[t]
   \centering
    \includegraphics[width = .99\linewidth]{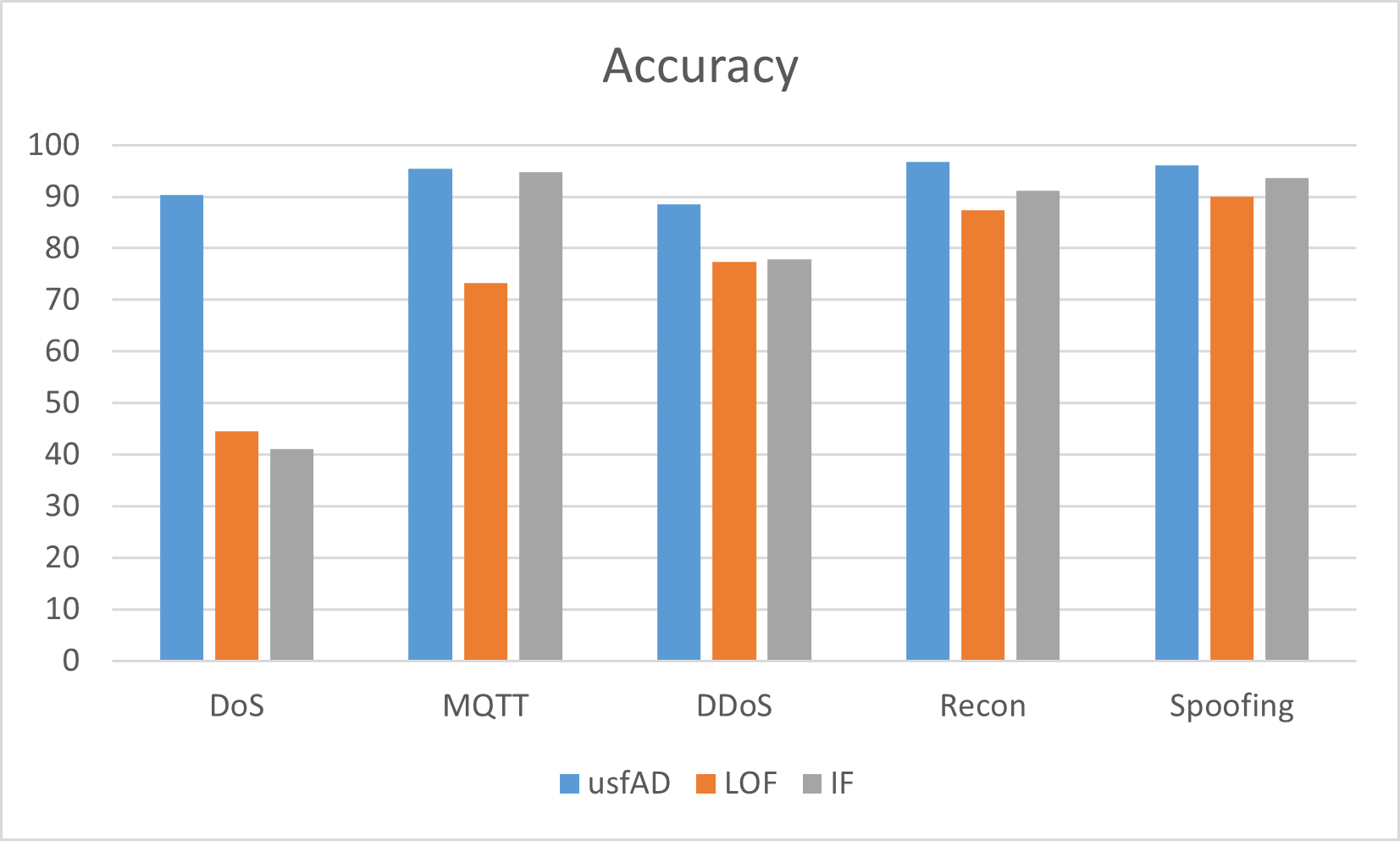}
    \caption{The accuracy of OCC to detect known and unknown attack types}
    \label{fig:accuracy2ndlevel}
\end{figure}

Figure \ref{fig:accuracy2ndlevel} demonstrates the accuracy of three one-class classifiers—usfAD, LOF, and IF—for detecting known and unknown attacks at the second level of a detection framework. 
    Specifically, usfAD consistently achieves the highest accuracy across all attack types, with notable performance for Recon (96.69\%) and Spoofing (96.01\%). 
    This indicates its robustness and ability to generalize effectively across different attack scenarios, making it the most reliable option for distinguishing between known and unknown attacks.

LOF shows significant variability, performing poorly for DoS (44.49\%) when DoS is unknown but achieving better accuracy for Spoofing (90.07\%) when it is unknown. 
    IF performs better than LOF for most attack types, with good accuracy for MQTT (94.81\%) and Spoofing (93.7\%), but falls behind usfAD overall. 
    However, both LOF and IF exhibit inconsistencies, highlighting their limited reliability for comprehensive detection tasks. 
    Overall, usfAD stands out as the most robust and accurate classifier for detecting both known and unknown attacks.

\subsection{Refined Attack Classifications at the Top Levels}

 Recall that in our framework, the first level filters attack instances and passing them to the next level. 
    Then, the second level OCC is trained to detect known and known attack.
    Finally, the third and fourth levels (which is deployed on the Cloud) will further examine the suspicious traffic to identify its attack category and sub-category, respectively.
    For these top two levels, our framework leverages Random Forest (RF) due to its effectiveness in classifying IoMT attacks \cite{dadkhah2024ciciomt2024, ramesh2024efficient}.    
It is worth noting that some normal instances are misclassified as attacks by the OCC at the root/first level, and then it will be passed to the second layer.
    Similarly, if misclassified normal instance is detected as known, it is passed to third and fourth levels.
    Therefore, at the third and fourth levels, we trained RF models using both normal and attack instances, namely RF1 models.
    By doing so, these RF models can correct the misclassified instances that were mistakenly forwarded by the previous layer. 

\begin{table*}[t]
    \caption{Performance metrics for Random Forest models }
    \label{tab:performance_metrics}
    \centering
    \begin{tabular}{|p{2.5cm}|c|c|c|c|c|c|}
        \hline
        \multirow{2}{*}{\textbf{Category}} & 
        \multicolumn{3}{c|}{\textbf{RF1 (Benign + Attacks)}} & 
        \multicolumn{3}{c|}{\textbf{RF2 (Attacks Only)}} \\
        \cline{2-7}
         & \textbf{Precision} & \textbf{Recall} & \textbf{F1-Score} & \textbf{Precision} & \textbf{Recall} & \textbf{F1-Score} \\
        \hline
        Benign    & 88.73  & 98.56  & 93.39  & 0.00   & 0.00   & 0.00   \\
        DoS       & 99.97  & 99.97  & 99.97  & 99.97  & 99.97  & 99.97  \\
        MQTT      & 99.98  & 99.96  & 99.97  & 99.90  & 99.98  & 99.94  \\
        DDoS      & 99.99  & 99.99  & 99.99  & 99.99  & 99.99  & 99.99  \\
        Recon     & 99.89  & 99.75  & 99.82  & 97.38  & 99.89  & 98.62  \\
        Spoofing  & 98.32  & 96.20  & 97.25  & 90.99  & 97.12  & 93.96  \\
        \hline
        \textbf{Accuracy}  & \multicolumn{3}{c|}{\textbf{99.97}} & \multicolumn{3}{c|}{\textbf{99.93}} \\
        \hline
        Macro Avg & 97.81  & 99.07  & 98.40  & 81.37  & 82.82  & 82.08  \\
        Weighted Avg & 99.98  & 99.97  & 99.98  & 99.88  & 99.93  & 99.91  \\
        \hline
    \end{tabular}
\end{table*}

\begin{table*}[t]
    \centering
    \caption{Performance metrics for Random Forest models by subcategory}
    \label{tab:performance_metrics_subcategory}
    \begin{tabular}{|p{3.5cm}|c|c|c|c|c|c|}
        \hline
        \multirow{2}{*}{\textbf{Subcategory}} & 
        \multicolumn{3}{c|}{\textbf{RF1 (Benign + Attacks)}} & 
        \multicolumn{3}{c|}{\textbf{RF2 (Attacks Only)}} \\
        \cline{2-7}
         & \textbf{Precision} & \textbf{Recall} & \textbf{F1-Score} & \textbf{Precision} & \textbf{Recall} & \textbf{F1-Score} \\
        \hline
        Benign                 & 88.09  & 98.80  & 93.14  & 0.00   & 0.00   & 0.00   \\
        ARP\_Spoofing          & 98.37  & 96.35  & 97.35  & 86.31  & 98.20  & 91.87  \\
        Recon-Port\_Scan       & 94.97  & 98.45  & 96.68  & 94.51  & 98.50  & 96.46  \\
        MQTT-DoS-Publish\_Flood & 99.95  & 99.97  & 99.96  & 99.95  & 99.98  & 99.97  \\
        MQTT-DDoS-Publish\_Flood & 99.94  & 99.93  & 99.94  & 99.94  & 99.92  & 99.93  \\
        Recon-OS\_Scan         & 89.98  & 74.38  & 81.44  & 83.72  & 75.01  & 79.13  \\
        MQTT-DoS-Connect\_Flood & 100.00 & 99.85  & 99.92  & 99.97  & 99.85  & 99.91  \\
        MQTT-Malformed\_Data   & 98.43  & 98.32  & 98.38  & 93.07  & 99.11  & 95.99  \\
        Recon-VulScan          & 95.35  & 71.93  & 82.00  & 74.14  & 75.44  & 74.78  \\
        Recon-Ping\_Sweep      & 88.60  & 87.83  & 88.21  & 66.01  & 87.83  & 75.37  \\
        TCP\_IP-DDoS-ICMP      & 99.98  & 99.99  & 99.99  & 99.99  & 99.99  & 99.99  \\
        TCP\_IP-DDoS-TCP       & 99.99  & 99.98  & 99.98  & 99.99  & 99.98  & 99.99  \\
        TCP\_IP-DDoS-SYN       & 99.99  & 99.99  & 99.99  & 99.99  & 99.99  & 99.99  \\
        TCP\_IP-DoS-UDP        & 99.96  & 99.94  & 99.95  & 99.96  & 99.95  & 99.95  \\
        TCP\_IP-DoS-SYN        & 99.98  & 99.97  & 99.97  & 99.97  & 99.97  & 99.97  \\
        TCP\_IP-DoS-ICMP       & 99.96  & 99.93  & 99.95  & 99.96  & 99.94  & 99.95  \\
        TCP\_IP-DoS-TCP        & 99.95  & 99.98  & 99.97  & 99.95  & 99.99  & 99.97  \\
        TCP\_IP-DDoS-UDP       & 99.98  & 99.99  & 99.99  & 99.98  & 99.99  & 99.99  \\
        Recon-Port\_Scan       & 99.99  & 100.00 & 99.99  & 99.99  & 100.00 & 99.99  \\
        \hline
        \textbf{Accuracy}      & \multicolumn{3}{c|}{\textbf{99.89}} & \multicolumn{3}{c|}{\textbf{99.84}} \\
        \hline
        Macro Avg              & 97.55  & 96.08  & 96.67  & 89.34  & 91.24  & 90.17  \\
        Weighted Avg           & 99.88  & 99.89  & 99.88  & 99.80  & 99.84  & 99.82  \\
        \hline
    \end{tabular}
\end{table*}

In the literature, some studies (e.g.,\cite{uddin2024usfad}) have recommended training supervised classifiers solely on attack instances. 
    However, in their approaches, any misclassified normal instances at the first level remain undetected at the subsequent levels, thereby adversely affecting the overall accuracy of the model.
    Therefore, to evaluate and compare the effectiveness of the proposed approach (i.e., training RF with both normal and attack traffic), in the third and fourth levels, we also trained other RF models (namely RF2 models) using only different known attack categories.
   
The left side of Tables \ref{tab:performance_metrics} and \ref{tab:performance_metrics_subcategory} presents the performance of RF model trained using both normal and attack instances, i.e., RF1, while the right side of them presents the performance of the RF model trained only using attack instances, i.e., RF2.  
    These results indicate that the first level does not perfectly distinguish between normal and attack instances. 
    Specifically, approximately 831 normal instances were misclassified as attacks and forwarded to the second level.     
    Since RF2 models at the second and third levels were trained solely on attack instances, they failed to correctly identify these 831 misclassified normal instances.     
On the other hand, by training with both attack and normal instances, RF1 models can correctly identify normal instances that were erroneously passed from the first level.
    Consequently, this solution allows the second and third levels detect misclassifications forwarded by the first layer effectively. 


Given the above results, while the first level with meta-learning or OCC is essential for detecting emerging attacks, using supervised learning at the higher levels (i.e., three and four) enhances overall accuracy and allows for the identification and correction of misclassifications.
    Thus, our proposed hierarchical approach not only improves detection rates but also facilitates appropriate actions in response to potential misclassifications.

\subsection{Comparison of the Proposed Model with Existing Models}


\begin{table*}[t]
\caption{Performance comparison of state-of-the-art literature}
\label{tab:comparisonresult}
\begin{tabular}{|p{1.35cm}|p{1cm}|p{1cm}|p{1cm}|p{1.1cm}|p{2cm}|p{7cm}|}
\hline
\textbf{References} & \textbf{Accuracy} & \textbf{Recall} & \textbf{Precision} & \textbf{F1-score} & \textbf{Algorithm} & \textbf{Remarks} \\ \hline
Dadkhah et al.\cite{dadkhah2024ciciomt2024} & 99.6 & 95.1 & 97.1 & 96.1 & Random Forest & Developed CIC-IoMT2024 dataset. 

Did not address zero day attack issues. \\ \hline
Mohammadi et al.\cite{mohammadi2024securing} & 99 & 99 & 98 & 98 & CNN & Leveraged deep learning for detecting known attacks. 

Did not address zero day attack issues. \\ \hline
Minh et al.\cite{vu2024performance} & 99.56 & 91.28 & 94.59 & 92.62 & Swarm Learning & Distributed learning issue was addressed. 

Did not address zero day attack issues. \\ \hline
Sohail et al.\cite{sohail2024explainable} & 95.01 & - & - & - & XGBoost & Applied explainable boosting ensemble methods for multi-classification. 

Did not address zero day attack issues. \\ \hline
Ramesh et al.\cite{ramesh2024efficient} & - & 99.82 & 99.81 & 99.81 & Random Forest & Applied Recursive Feature Elimination with Cross Validation (RFECV). 

Did not address zero day attack issues. \\ \hline
{Proposed Approach-1} & 98.27 & 96.25 & 98.29 & 97.2 & Meta-learning Algorithm & Capable of identifying historical and zero-day attacks. 

Requires very few samples to train and retrain. \\ \hline
Proposed Approach-2 & 99.77 & 99 & 97.81 & 97.81 & usfAD (One Class Classifier) & Capable of identifying historical and zero-day attacks. 

Trained using normal instances only. \\ \hline
\end{tabular}
\end{table*}

The comparative analysis of the algorithms presented in Table \ref{tab:comparisonresult} highlights the strengths and limitations of state-of-the-art methods with our works for root level (i.e., binary classification-anomaly detection). 
    None of the existing studies specifically considered hierarchical IDSs and addressed zero-day attacks using the CIC-IoMT2024 dataset within IoMT networks. 
    Still, we compare our approach with these studies to demonstrate its advantages for detecting normal instances and zero day attacks which is overlooked by existing works in IoMT domain. 
    However, we do not provide a direct performance comparison for detecting known and unknown attacks at the second level, as these existing methods did not focus on detecting known and unknown attacks.

As shown in Table~\ref{tab:comparisonresult}, traditional machine learning models like Random Forest (used by Dadkhah et al. \cite{dadkhah2024ciciomt2024} and Ramesh et al. \cite{ramesh2024efficient}) show high accuracy and precision but fail to detect zero-day attacks because they rely on supervised learning, which requires training on specific attack instances. 
    Similarly, CNN, used by Mohammadi et al. \cite{mohammadi2024securing}, also performs well but lacks the capability to detect zero-day attacks, highlighting the limitations of deep learning models trained exclusively on known data. 
    Swarm Learning (Minh et al. \cite{vu2024performance}) and XGBoost (Sohail et al. \cite{sohail2024explainable}) are effective for distributed learning and multi-class classification, but like other methods, they do not address zero-day attacks.

In contrast, our proposed approaches, i.e., Meta-learning Classifier (MLC) (Approach-1) and usfAD (Approach-2), stand out due to their ability to detect both historical and zero-day attacks. 
    The meta-learning classifier (MLC) achieves this by requiring only a few samples for training, making it particularly suitable for scenarios with limited data. 
    Similarly, usfAD, a one-class classifier, detects unknown attacks by training exclusively on normal instances, making it robust in handling anomalies, including zero-day threats. 
    Our experimental results demonstrate that our proposed approaches can achieve comparable or even better detection performance than traditional methods, despite the additional challenge of handling zero-day attacks that other existing approaches~\cite{dadkhah2024ciciomt2024, ramesh2024efficient, mohammadi2024securing, vu2024performance, sohail2024explainable} do not consider.
    These approaches showcase the evolving capabilities of attack detection, offering a significant advantage over traditional methods by requiring minimal training data and providing the ability to generalize to unknown attack types while yet achieving a high accuracy.



\section{Conclusions and Future Work}
\label{Conclusion}

In this paper, we have analyzed the current Intrusion Detection Systems (IDS) in the Internet of Medical Things (IoMT), highlighting significant limitations in conventional flat and centralized multi-class classification approaches.
    These models fail to address the dynamic and distributed nature of IoMT networks and are ineffective in detecting zero-day attacks due to their reliance on historical data. 
    To overcome these challenges, we propose a hierarchical IDS that strategically distributes various classifiers on different layers of IoMT's hierarchical networks, offering a robust and adaptive solution.
    The proposed system leverages meta-learning or usfAD algorithms at the root level to detect zero-day attacks with minimal training data or even without requiring attack instances, making it highly efficient and adaptable for deployment in resource-constrained environments such as medical end devices. 
    At subsequent levels of the hierarchy, the system incorporates one-class and supervised classifiers to determine whether the attack's type is known and identify its specific subcategory. 
    This approach enables our system to provide detailed and actionable insights, ensuring comprehensive defense against evolving cyber threats
    Moreover, the proposed approach not only enhances intrusion detection accuracy but also aligns with the computational and operational constraints of modern IoMT networks, paving the way for more secure and efficient network infrastructures.
    Therefore, our proposed hierarchical architecture can effectively mitigate the critical challenges of scalability, communication overhead, and real-time decision-making in distributed networks like IoMT.




\bibliographystyle{IEEEtran}
\bibliography{references}



\newpage

 




\vfill

\end{document}